\documentclass[journal,10pt]{IEEEtran}
\usepackage{cite}
\usepackage{graphicx}
\usepackage{epstopdf}
\usepackage{amsmath}
\usepackage{times}
\usepackage{latexsym}
\usepackage{graphicx}
\usepackage{bm}
\usepackage{bbm}
\usepackage{amssymb}
\usepackage[center]{caption2}
\usepackage{stfloats}
\usepackage{cases}
\usepackage{array}
\usepackage{setspace}
\usepackage{fancyhdr}
\usepackage{fancyhdr}
\usepackage{algorithm}
\usepackage{algorithmic}
\usepackage{textcomp}
\usepackage{wasysym}
\usepackage{balance}
\usepackage{flushend}
\usepackage{url}
\usepackage{xcolor}

\graphicspath{{figures/}}

\allowdisplaybreaks[4]

\newcaptionstyle{mystyle1}{%
  \centering TABLE  \captiontext \par}
\captionstyle{mystyle1}
\newcaptionstyle{mystyle2}{%
  \captionlabel $.$ \: \captiontext \par}
\captionstyle{mystyle2}
\newcaptionstyle{mystyle3}{%
  \captionlabel $.$ \: \captiontext \par}
\captionstyle{mystyle3}

\begin{document}

\title{ Distributed Edge Caching via Reinforcement \\ Learning in Fog Radio Access Networks}


\author{Liuyang~Lu,
Yanxiang~Jiang,~\IEEEmembership{Senior~Member,~IEEE},
Mehdi~Bennis,~\IEEEmembership{Senior~Member,~IEEE}, 
Zhiguo~Ding,~\IEEEmembership{Senior~Member,~IEEE}, 
Fu-Chun~Zheng,~\IEEEmembership{Senior~Member,~IEEE},
and Xiaohu~You,~\IEEEmembership{Fellow,~IEEE}
\thanks{This work has been accepted by IEEE VTC 2019 Spring.}
\thanks{L. Lu and Y. Jiang are with the National Mobile Communications Research Laboratory, Southeast University, Nanjing 210096, China,
the State Key Laboratory of Integrated Services Networks, Xidian University, Xi'an 710071, China,
and the Key Laboratory of Wireless Sensor Network $\&$ Communication, Shanghai Institute of Microsystem and Information Technology,
Chinese Academy of Sciences, Shanghai 200050, China. (e-mail: yxjiang@seu.edu.cn)}
\thanks{M. Bennis is with the Centre for Wireless Communications, University of Oulu, Oulu 90014, Finland. (e-mail: bennis@ee.oulu.fi)}
\thanks{Z. Ding is with the School of Electrical and Electronic Engineering, University of Manchester, Manchester, UK. (e-mail: zhiguo.ding@manchester.ac.uk)}
\thanks{F. Zheng is with the School of Electronic and Information Engineering, Harbin Institute of Technology, Shenzhen 518055, China, and the National Mobile Communications Research Laboratory, Southeast University, Nanjing 210096, China. (e-mail: fzheng@ieee.org)}
\thanks{X. You is with the National Mobile Communications Research Laboratory, Southeast University, Nanjing 210096, China. (e-mail: xhyu@seu.edu.cn)}
}

\maketitle
\begin{abstract}
In this paper, the distributed edge caching problem in fog radio access networks (F-RANs) is investigated. By considering the unknown spatio-temporal content popularity  and user preference, a user request model based on hidden Markov process is proposed to characterize the fluctuant spatio-temporal traffic demands in F-RANs. Then, the Q-learning method based on the reinforcement learning (RL) framework is put forth to seek the optimal caching policy in a distributed manner, which enables  fog access points (F-APs) to learn and track the potential dynamic process without extra communications cost. Furthermore, we propose a more efficient Q-learning method with value function approximation (Q-VFA-learning) to reduce  complexity and accelerate convergence.
Simulation results show that the performance of our proposed method is superior to those of the traditional methods.
\end{abstract}

\section{Introduction}

The rapid development of smart devices and mobile applications brings an unprecedented traffic pressure to the wireless networks \cite{E.A}.
To cope with this challenge and meet the stringent quality of service (QoS) standards, significant changes to the cellular infrastructure are required\cite{cisco}.
One promising such change is through the dense deployment of caching resources at   network edges\cite{edgee}.
At this point, fog radio access network (F-RAN) \cite{F-RAN} has been proposed as an evolution form of heterogeneous cloud radio access networks.
By taking full advantage of computing and storage capabilities in edge devices, F-RANs can effectively reduce the backhaul load by prefetching and caching the most popular contents.
In F-RAN, edge devices with limited resources are fog access points (F-APs).
Due to the resource constraints, distributed caching has been considered as a practical mechanism in F-RANs, which does not require information exchange between neighboring F-APs, thus can reduce the network overhead effectively.
To strategically prefetch contents in a distributed manner, each F-AP must learn how, what and when to cache, while taking into account storage limitations, cache refreshing costs and fluctuant spatio-temporal traffic demands.

There are many traditional distributed caching methods including belief propagation-based algorithm \cite{bf}, alternation direction method of multipliers algorithm \cite{admm}, and game theory based algorithm \cite{yabai}.
However, those existing works \cite{bf, admm, meansquare} presume that the popularity profile is known in advance and keeps stationary over long time span, which is not practical.
Hence, an increasing number of works in distributed caching have focused on entailing the edge ability of learning content popularity.
In \cite{learfile}, a perturbed version of content popularity was learned in an online fashion to simplify the decentralized cache problem to a multi-armed bandit problem.
The cache content placement problem in single base station was cast into a reinforcement learning (RL) framework and an online method to learn the spatial and temporal content popularity profile was presented in \cite{pupl}.
This method considers the influence of the global popularity towards the requests of a local region, which inevitablely introduces additional information exchanging cost.
To reduce this kind of network overhead, a distributed caching policy based on a game of independent learning automata was proposed in \cite{dis-len} to minimize the downloading latency by observing the instantaneous exchanged information between learners and environment.
This policy can  achieve a fully distributed caching procedure which ignores the priori knowledge of regional content popularity, whereas it leads to a slow learning speed.
However, the content popularity considered in \cite{learfile, pupl, dis-len} is not prudent enough to reflect the demand statistic of users at a certain time and region.
Specifically, when there are few users with high mobility in the region, the probability of a particular content being requested is much more likely to depend on the long-term user characteristics \cite{user_p, user2}, while content popularity is more inclined to capture regional features.

Motivated by the aforementioned discussions and considerations, a distributed edge caching method is proposed in this paper to optimize the edge caching policy in F-RANs.
First, we propose a user request model to describe the fluctuant spatio-temporal traffic demands in F-RANs by considering the joint influence of content popularity and user preference.
Then, we cast the edge caching problem of each F-AP into the RL framework by defining a learning environment and designing a suitable reward function.
Furthermore, the Q-learning method is put forth to seek the optimal edge caching policy in a distributed manner, where each F-AP acts independently without extra information exchange.
Finally, we propose a more efficient Q-learning method with value function approximation (Q-VFA-learning) to reduce complexity and accelerate convergence.

The rest of this paper is organized as follows.
In Section II, the system model is elaborately described. The problem formulation and the proposed RL-based distributed edge caching method are presented in Section III.
Simulation results are provided in Section IV, and the main conclusions are drawn in Section V.

\section{System Model}

\subsection{Network Model}
\begin{figure}[!t]
\centering
\includegraphics[width=0.45 \textwidth]{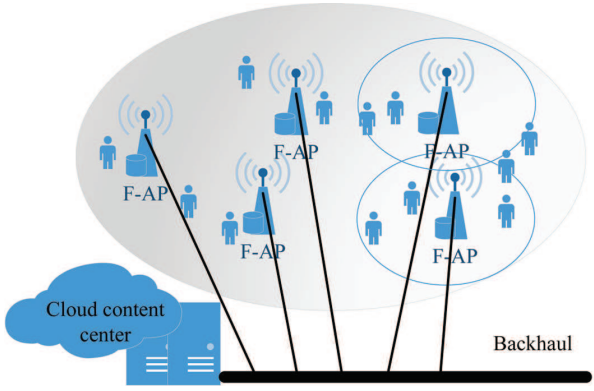}
\caption{Illustration of the edge caching scenario in the F-~RAN.}
\label{net}
\end{figure}

The considered F-RAN is illustrated in Fig. \ref{net}, where a large amount of F-APs are deployed.
At the network edge, the F-APs with limited resources are connected to the cloud content center  via the backhaul link.
Let $\mathcal F=\{1, 2, \ldots, f, \ldots, F\}$ denote the content library, which is located in the cloud content center.
For simplicity, it is assumed that all  contents have the exactly same size and each F-AP can store up to $B$ ($B\leq F$) contents.
Meanwhile, a time-slotted system  is considered.
Let $g_t$ denote  the user group that consists of  the users served by the F-AP during time slot $t$.
Assume that there are $N$ users in the user group $g_t$ and let $\mathcal N=\{1,2,\ldots,n,\ldots,N\}$ denote the set of the $N$ users.

\subsection{User Request Model}

The probability that a certain user requests some certain content is determined by  the spatio-temporal content popularity as well as the  user preference.
The content popularity and the user preference are modeled by using  two independent and identically distributed Markov chains as  shown in Fig. \ref{mdp}, where the corresponding information is collected at the beginning of each time slot.
This kind of setting characterizes an environment with   fluctuant spatio-temporal traffic demands.
In the setting, the user group $g_t$  consists of the dynamically arriving and leaving users, while a certain user's preference  is  fixed.
Meanwhile, the content popularity is variant over time and space.
Let $\cal{P}$ $=\left\{ {\boldsymbol{P}_1,\boldsymbol{P}_2,\ldots,\boldsymbol{P}_{|\cal{P}|}} \right\}$ and $\cal{Q}$ $=\left\{ {\boldsymbol{Q}_1,\boldsymbol{Q}_2,\ldots,\boldsymbol{Q}_{|\cal{Q}|}} \right\}$
denote the implied state  sets of the content popularity and the user preference.
Specifically, the dimensions of the implied states in the Markov chains are assumed known in advance, while the underlying transition probabilities are considered unknown.

\subsubsection{Content popularity}

Let $d_f(t)$ denote the amount of  the instantaneous user requests towards the $f$th content during time slot $t$.
Correspondingly, Let $p_f(t)$ denote the regional content popularity of the $f$th content,
which can be expressed as follows,
\begin{eqnarray}
{{p}_{f}(t)=\frac{d_f(t)}{\sum \nolimits_{f \in \cal F}{d_f(t)}}}.
\end{eqnarray}

\subsubsection{User preference}

During time slot $t$, let ${\boldsymbol{x}_{n}(t)} \in \mathbb{R}^M$ denote the user characteristic vector \cite{user} with dimension $M$ describing the preferences of the $n$th user.
Meanwhile, let ${\boldsymbol{y}_{f}(t)} \in \mathbb{R}^M$ denote the content feature vector \cite{validity} with dimension $M$ describing the features of the $f$th content.
Take video as an example: the user characteristics and content features  can be described in the aspects such as time validity, video type, video resolution.
Without loss of generality, we normalize the various dimensions of ${\boldsymbol{x}_{n}(t)}\in [0,1]^M$ and ${\boldsymbol{y}_{f}(t)}\in [0,1]^M$.
Then, we  introduce a parameter $\alpha$  to the  kernel function in \cite{kernel} to  dynamically reflect the correlation between the $n$th user and  the $f$th content.
Let $g[{\boldsymbol{x}_{n}(t)},{\boldsymbol{y}_{f}(t)}]$ denote the kernel function.
Then it  can be expressed as follows,
\begin{eqnarray}
{g[{\boldsymbol{x}_{n}(t)},{\boldsymbol{y}_{f}(t)}]={{(1-<{\boldsymbol{x}_{n}(t)},{\boldsymbol{y}_{f}(t)}>)}^{\log (1-\alpha )}}} \label{xy},
\end{eqnarray}
where $<,>$ denotes the   inner product operator.
Note that $0\le \alpha <1$.
When $\alpha \to {{1}^{-}}$, we have,
\begin{align}
{g[{\boldsymbol{x}_{n}(t)},{\boldsymbol{y}_{f}(t)}]\to \left\{ {\begin{array}{*{20}{l}}
0,\text{   }{\boldsymbol{x}_{n}(t)}\ne {\boldsymbol{y}_{f}(t)}\text{ } \\
1,\text{   }{\boldsymbol{x}_{n}(t)}={\boldsymbol{y}_{f}(t)}, \\
\end{array}} \right.}
\end{align}
which indicates that no user has the same preference.
When $\alpha=0$, we have $g[{\boldsymbol{x}_{n}(t)},{\boldsymbol{y}_{f}(t)}]=1$ for any $n$ and $f$,
which indicates that the user preferences are the same for every content.
According to \eqref{xy}, we can deduce that $g[{\boldsymbol{x}_{n}(t)},{\boldsymbol{y}_{f}(t)}]$ takes values in $[0,1]$, and a lower value indicates a higher probability of the corresponding request.
Let ${q}_{f}(t)$ denote the user preference of the user group $g_t$ during time slot $t$ for the $f$th content, which can be expressed as follows,
\begin{eqnarray}
{q_{f}{(t)}=\frac{1}{N}{\sum\nolimits_{n\in \cal{N}}{g[\boldsymbol{x}_{n}(t),\boldsymbol{y}_{f}(t)]}}}.
\end{eqnarray}

\subsection{Edge Caching Model}
\begin{figure}[!t]
\centering
\includegraphics[width=0.45 \textwidth]{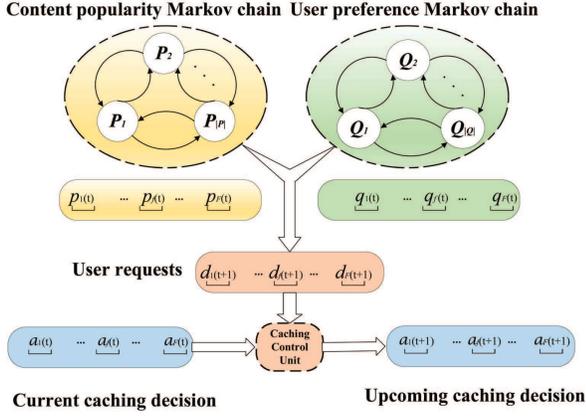}
\caption{The schematic depicting the user request model and the edge caching model.}
\label{mdp}
\end{figure}

The considered  time-slotted system is depicted in Fig. \ref{mdp} where a batch of requests arrives at the beginning of each time slot $t = 1,2,\ldots,T$, where $T$ is considered to be a finite time horizon.
{Let $a_{f}(t)$ denote the cache indicator for the considered F-AP during time slot $t$.
Specifically, $a_{f}(t)=1$ indicates that the $f$th content is cached during time slot $t$ and $a_{f}(t)=0$ otherwise.
Correspondingly, caching decisions are made in the control unit of the considered F-AP.
At the end of  each time slot $t$, the control unit  changes the current caching indicator $a_{f}(t)$ to the upcoming  $a_{f}(t+1)$.
}
\section{The RL-Based distributed Edge Caching Method}
In this section, we first formulate the distributed edge caching optimization problem in F-RANs by casting the problem into the RL framework.
Then, the   Q-VFA-learning method is proposed to seek the optimal edge caching policy.

 \subsection{Problem Formulation   in the RL Framework}
Distributed edge caching effectively reduces the   communication overhead since  F-APs are supposed to act independently without extra information exchange.
In the RL framework, the RL agent only uses the evaluative feedback from the environment as a performance measure.
Hence RL is a proper tool to realize the sequential decision problem of the distributed edge caching.
During time slot $t$, the RL agent receives a state $\boldsymbol{s}(t)$ in the state space $\cal{S}$ and selects an action $\boldsymbol{a}(t)$ from the action space $\cal{A}$ with probability $P(\boldsymbol{s}'|\boldsymbol{s},\boldsymbol{a})=\mathbb{P}[\boldsymbol{s}(t+1)=\boldsymbol{s}'|\boldsymbol{s}(t)=\boldsymbol{s},\boldsymbol{a}(t)=\boldsymbol{a}]$, where $\sum\nolimits_{\boldsymbol{s}' \in \cal{S} }P(\boldsymbol{s}'|\boldsymbol{s},\boldsymbol{a})=1$.
The agent's behavior here is a policy $\pi(\boldsymbol{a}|\boldsymbol{s})$, which is also considered as a mapping from state $\boldsymbol{s}$ to action $\boldsymbol{a}$.
The instantaneous reward of taking action $\boldsymbol{a}(t)$ in state $\boldsymbol{s}(t)$ is $r(t)$.
The object of the agent is to find a policy, that minimizes the long-term reward value from each state\cite{reinforcement}.

To eventually formulate the problem in the RL framework,   the four fundamental elements in the caching optimization problem are to be introduced.
Specifically, the F-AP is the learning agent in this problem.

\emph{State Space:}
 $\forall f \in \cal F$, let the user preference $q_{f}(t)$,  content popularity $p_{f}(t)$ and the last cached decision $a_{f}(t-1)$ form the state space of the agent F-AP during time slot $t$.
For simplification, let  $\boldsymbol{p} =[ p_{1}(t),\ldots,p_{f}(t),\ldots,p_{F}(t) ]^T$ and $\boldsymbol{q} =[ q_{1}(t),\ldots,q_{f},\ldots,q_{F}(t) ]^T$ denote the current content popularity vector and the current user preference vector.
Let $\cal{S}$ $=\left\{ {\boldsymbol{s}_1,\boldsymbol{s}_2,\ldots,\boldsymbol{s}_{|\cal{S}|}} \right\}$  denote the state space, where $|S|$ represents the dimension  of the state space.

\emph{Action Space:}
In order to meet the storage constraint of the F-AP, every possible action in the action set $\cal{A}$ should satisfy the following condition,
\begin{eqnarray}
{\sum\nolimits_{f \in \cal F }a_{f}(t)=B, a_{f}(t)\in \{0,1\}}.
\end{eqnarray}
Note that the dimension of the action set $\cal{A}$ is $|\cal{A}|$.
Let $\boldsymbol{a}(t) = [ a_{1}(t),\ldots,a_{f}(t),\ldots,a_{F}(t) ]^T$ denote the current action, likewise the upcoming action $\boldsymbol{a}(t+1)$.

\emph{Reward:}
Cache hit rate is a common measure of the caching performance,  hence it is adopted as  the reward function in the RL framework for this optimization problem.
Let $\theta(t)$   denote the cache hit rate during time slot $t$,
which can be expressed as follows,
\begin{eqnarray}
{\theta(t)=\frac{\sum\nolimits_{f \in \cal F }d_f(t)a_f(t)}{\sum\nolimits_{f \in \cal F }d_f(t)}}.
\end{eqnarray}
Let $r(t)$ denote the instantaneous reward during time slot $t$, which can be expressed as follows,
\begin{eqnarray}
{r(t)=1-\theta(t)}\label{r}.
\end{eqnarray}
The agent can achieve a good performance of the long-term cache hit rate when the average   reward is minimized.

\emph{Action Value Function:}
Let $Q_\pi(\boldsymbol{s},\boldsymbol{a})$ denote the value function following the policy $\pi$.
Then, it can be expressed as follows,
\begin{eqnarray}
{Q_\pi(\boldsymbol{s},\boldsymbol{a})=\mathit{E}[\sum\nolimits_{\tau=t }^{\infty}{\gamma }^{\tau -t}r(\boldsymbol{s}(\tau),\boldsymbol{a}(\tau))]_{\boldsymbol{s}(\tau)=\boldsymbol{s},\boldsymbol{a}(\tau)=\boldsymbol{a}}},\notag
\end{eqnarray}
where $\gamma \in \text{(0,1]}$ denotes the discount factor.
The discount factor $\gamma$ reflects to what degree the future reward is affected by the past actions.
Action value function is a prediction of the expected accumulative reward over infinite time horizon, which measures the optimality of  each state-action pair.
In other words, action value function helps to determine which and when the content should be cached.
Due to the Markov property, i.e., the state at the subsequent time slot is only determined by the current state and irrelevant to the former states, the action value function can be decomposed into the Bellman equation
$Q_\pi(\boldsymbol{s},\boldsymbol{a})=\sum\nolimits_{\boldsymbol{s}'}p(\boldsymbol{s}'|\boldsymbol{s},\boldsymbol{a})[r+\gamma \sum\nolimits_{\boldsymbol{a}'}\pi(\boldsymbol{a}'|\boldsymbol{s}')Q_{\pi}(\boldsymbol{s}',\boldsymbol{a}')]$.
The optimal action value function  $Q^*(\boldsymbol{s},\boldsymbol{a})=\min \nolimits_\pi Q_\pi(\boldsymbol{s},\boldsymbol{a})$ can be decomposed into the Bellman optimality equation $Q^*(\boldsymbol{s},\boldsymbol{a})=\sum\nolimits_{\boldsymbol{s}'}p(\boldsymbol{s}'|\boldsymbol{s},\boldsymbol{a})[r+\gamma \min \nolimits_{\boldsymbol{a}'} Q^*(\boldsymbol{s}',\boldsymbol{a}') ]$.
Exploiting the Bellman equation and the Bellman optimality equation, we can use RL to solve the dynamic programming problem\cite{reinforcement}.
First, the agent  evaluates   $Q_\pi(\boldsymbol{s},\boldsymbol{a})$ for the current policy $\pi$.
Then, the policy is updated as follows,
\begin{eqnarray}
{\pi'=\arg \min \limits_{\boldsymbol{a}}{Q}_\pi(\boldsymbol{s},\boldsymbol{a})}\notag.
 \end{eqnarray}

The objective of this paper is to determine the optimal policy $\pi^*$ such that the average reward of any state $\boldsymbol{s}$ can be minimized.
More specifically, the RL-based optimization problem is formulated as follows,

\begin{eqnarray}
{\min \limits_\mathbf{\pi^*}\quad\mathit{E}[Q_{\pi^*}(\boldsymbol{s},\boldsymbol{a})], \forall \boldsymbol{s} \in \cal{S}}.
\end{eqnarray}

\subsection{Optimal Edge Caching via Q-learning Method and Q-VFA-learning Method}
\subsubsection{ Q-learning Method}
Q-learning is an off-policy control method and an online algorithm in the  RL framework, which is one of the most widely-used strategies to determine the best policy $\pi^*$.
By considering the unknown  information of $P(\boldsymbol{s}'|\boldsymbol{s},\boldsymbol{a})$, $\forall\boldsymbol{s},\boldsymbol{a}$, the Q-learning method is a typical  adaptive dynamic programming method \cite{adp} that learns and tracks the potential dynamic environment, without the need to estimate   $P(\boldsymbol{s}'|\boldsymbol{s},\boldsymbol{a})$, $\forall\boldsymbol{s},\boldsymbol{a}$.
The  update rule can be expressed as follows,
\begin{eqnarray}
{Q(\boldsymbol{s},\boldsymbol{a})\gets Q(\boldsymbol{s},\boldsymbol{a})+\delta_t[r+\gamma \min\limits_{\boldsymbol{a}'} Q(\boldsymbol{s'},\boldsymbol{a'}) -Q(\boldsymbol{s},\boldsymbol{a})]},
\end{eqnarray}
where $\delta_t$ is the learning rate.
When the agent observes the environment during the initial iterations,  $\delta_t$ should be set relatively large for the agility of the learning process.
While enough observations have been accumulated, $\delta_t$ should be set to a small value to maintain the precision and reliability of the learning process.
Hence, we set the learning rate $\delta_t=\frac{1}{\sqrt{t+2}}$ as  a function of time.
Meanwhile, a probabilistic exploration-exploitation approach is utilized to determine the future actions, which guarantees the convergence of the Q-learning method.
\subsubsection{ Q-VFA-learning Method}
The original Q-learning method saves the action value function $Q_\pi(\boldsymbol{s},\boldsymbol{a})$ in tabular form.
Despite the simple and comprehensive features,  the applicability of Q-learning in tabular form  faces practical challenges in real networks, since the actual state or action space is large or continuous.
Value function approximation  \cite{reinforcement} can cope with the case that the state space is unsuited for explicit representation.
With a proper function approximation model, the agent can estimate the state value in the partitioned space that has been visited,   induce the value in cross-region \cite{reinforcement}, and then directly estimate the value in the corresponding space without requiring every continuous state.
Linear value function approximation \cite{VFA} is a way to generalize the  Q-learning method  in real setting, which is named as Q-VFA-learning method in the following.

We propose a value function approximation model, which considers  the induced backhaul load,  the mismatch cost between the caching decision and the content popularity as well as the consideration of satisfying the user preference.

Let ${z}_{1}(\boldsymbol{s},\boldsymbol{a})$ denote the induced backhaul load  when the agent refreshes  the former caching decision to a new one.
Then, it can be formulated as follows,
\begin{eqnarray}
{{z}_{1}}(\boldsymbol{s},\boldsymbol{a})={{\boldsymbol{a}}^{T}}(1-\bar{\boldsymbol{a}})\label{3},
\end{eqnarray}
where $\bar{\boldsymbol{a}}$ is the original caching state during the current time slot.
The corresponding content in this case is cached during the previous time slot  while replaced by a new content.
According to the network model, the new content is sent to the F-AP over the backhaul link from the cloud content center.
Therefore, this type of cost represents the induced backhaul load.

Let $\boldsymbol{z}_{2}(\boldsymbol{s},\boldsymbol{a})$ denote the mismatch cost between the caching decision  and  the local content popularity during time slot $t$.
Then, it can be formulated as follows,
\begin{eqnarray}
{{\boldsymbol{z}_{2}(\boldsymbol{s},\boldsymbol{a})=({1-\boldsymbol{a}})\circ {{\boldsymbol{p}}}}}\label{2},
\end{eqnarray}
where  $\circ$ denotes the Hadamard product.
This cost is incurred when the content with high popularity is uncached, which    should be avoided.
Actually, it is reasonable to consider that the local content popularity indicates  the future requests in the region.
The external factors, such as the type of the device  and the location of F-APs, would  have an impact on users' demands, which are captured as the local content popularity.
For instance, in subway or commercial district, users prefer to requesting  short-form videos for entertainment due to the  traffic and battery limits.
While in living quarters, they have more kinds of alternative choices.

Let $\boldsymbol{z}_{3}(\boldsymbol{s},\boldsymbol{a})$ denote the degree of satisfying user preference  during time slot $t$, which can be formulated as follows,
\begin{eqnarray}
{{\boldsymbol{z}}_{3}}(\boldsymbol{s},\boldsymbol{a})=({1-\boldsymbol{a}})\circ {{\boldsymbol{q}}}\label{1}.
\end{eqnarray}
The   corresponding content in this case is not cached but the user preference collected at the moment is relatively higher than those cached.
Minimizing this cost  is beneficial to improve the prudence of the caching strategy  by considering user preference.
Specifically, when there are few users with high mobility in the region, the probability of a particular content being requested is much more likely to depend on the long-term user characteristic, while the content popularity is more inclined to capture regional features.

Given the three costs discussed above,  the cost vector with dimension $(2F+1)$  can be denoted as $\boldsymbol{z}(\boldsymbol{s},\boldsymbol{a})$, which can be expressed as follows,
\begin{eqnarray}
{\boldsymbol{z}(\boldsymbol{s},\boldsymbol{a})=[{z}_{1}(\boldsymbol{s},\boldsymbol{a}), \boldsymbol{z}_{2}(\boldsymbol{s},\boldsymbol{a})^T ,\boldsymbol{z}_{3}(\boldsymbol{s},\boldsymbol{a})^T]^T}.
\end{eqnarray}
The cost vector  considers  the joint influence of the backhaul   resource, the content popularity and the user preference.
Instead of simple superposition of the different induced costs
described in \cite{pupl}, here we consider more
specific influence of every content.
We propose to train the weight vector
$\boldsymbol{w}$ to reflect the relative importance of every content by
defining the approximate value function $\hat{Q}(\boldsymbol{s},\boldsymbol{a};\boldsymbol{w})$.
Then, it can be expressed as follows,
\begin{eqnarray}
{\hat{Q}(\boldsymbol{s},\boldsymbol{a};\boldsymbol{w})=\boldsymbol{z}(\boldsymbol{s},\boldsymbol{a})^T\boldsymbol{w}}.
\end{eqnarray}
According to  the approximate value function $\hat{Q}(\boldsymbol{s},\boldsymbol{a};\boldsymbol{w})$, we can determine the next action  which can minimize the costs.
In order to guarantee convergence,  a probabilistic exploration-exploitation approach is utilized to determine the future actions.

Given    the approximate value function $\hat{Q}(\boldsymbol{s},\boldsymbol{a};\boldsymbol{w})$ introduced above, here we present how to update the weight vector $\boldsymbol{w}$.
First, let  $\hat{\varepsilon} (\boldsymbol{s},\boldsymbol{a})$ denote the instantaneous error during the time slot $t$, which can be expressed as follows,
\begin{eqnarray}\label{vare}
{ \hat{\varepsilon } (\boldsymbol{s},\boldsymbol{a})=[r(\boldsymbol{s},\boldsymbol{a})+\gamma\underset{\boldsymbol{a}'}{\mathop{\min }}\,Q(\boldsymbol{s}',\boldsymbol{a}';{{\boldsymbol{w}}})-Q(\boldsymbol{s},\boldsymbol{a};{\boldsymbol{w}}){{]}^2}}.
\end{eqnarray}
Then, the weight vector $\boldsymbol{w}$ can be updated by using stochastic gradient descent method   \cite{reinforcement} as follows,
\begin{eqnarray}\label{w}
{\boldsymbol{w}} \gets {{\boldsymbol{w}}}+\rho \sqrt{\hat{\varepsilon} (\boldsymbol{s},\boldsymbol{a})} \boldsymbol{z}(\boldsymbol{s},\boldsymbol{a}),
\end{eqnarray}
where  $\rho$ is the step size.
The pseudo code for Q-VFA-learning method is presented in Algorithm 1, where parameter $\epsilon _{t}$ trades off exploration for exploitation during time slot $t$.
\subsubsection{Complexity Analysis}
The complexity of the policy evaluation step of Q-learning method  is $O(|\cal{S}||\cal{A}|)$, because the Q values \cite{lip} are updated for per state-action pair.
Furthermore, given ${Q}(\boldsymbol{s},\boldsymbol{a})$, $\forall$$\boldsymbol{s},\boldsymbol{a}$, the complexity of the policy update step is $O(|\cal{A}|)$.
Thus, the  complexity of Q-learning per iteration is $O(|\cal{S}||\cal{A}|)$.
In Q-VFA-learning, the complexity of the policy evaluation is $O({|\cal{A}|} |F|)$, and the complexity of the  weight vector update is $O(|F|)$.
Notice that ${|\cal{A}|}=\mathrm{C}_F^B$, which means that ${|\cal{A}|} {\gg F}$.
Therefore,  to a certain extent, the Q-VFA-learning method  could resolve the  curse of dimension  problem in traditional Q-learning method.
{\small{
\begin{algorithm}[!t]

	\caption{The Q-VFA-learning  based edge caching method}
	\label{alg:2}

	\begin{algorithmic}[1]
        \STATE Initialize $\boldsymbol{s}(0)$ and $\boldsymbol{a}(0)$ randomly, set $w(0)$ to 0;
		\FOR{$t = 1,2,\ldots$,$T$ }
                \STATE Record the previous action $\boldsymbol{a}(t-1)$;
            	\STATE Take the action $\boldsymbol{a}(t)$ chosen probabilistically by:
                \STATE \begin{displaymath}
                \boldsymbol{a}(t)=\left\{\begin{array}{ll}
   \arg \min \limits_{\boldsymbol{a}(t)}   \hat{Q}(\boldsymbol{s}(t),\boldsymbol{a}(t);\boldsymbol{w}) & \textrm{w.r.t. 1-${{\epsilon }_{t}}$} \\
   \textrm{random}\text{  }\boldsymbol{a}\in \cal A & \textrm{w.r.t. ${{\epsilon }_{t}}$} \\
\end{array} \right.
\end{displaymath}
                \STATE Observe the user preference $\boldsymbol{q}(t)$, the content popularity $\boldsymbol{p}(t)$;

                \STATE Determine the present state $\boldsymbol{s}(t)$;
                \STATE Calculate the reward $r(\boldsymbol{s},\boldsymbol{a})$ based on Eq. \eqref{r};
                \STATE Get the instantaneous error $\hat{\varepsilon } (\boldsymbol{s},\boldsymbol{a})$ based on Eq. \eqref{vare};
                \STATE Update the weight vector ${{\boldsymbol{w}}_{t}}$ based on Eq. \eqref{w}.
               		
        \ENDFOR

	\end{algorithmic}
\end{algorithm}
}}

\section{Simulation Results}
In this section, the performance of the proposed  RL-based edge caching method is evaluated.
We consider the  F-RAN with $F=20$   and the F-AP with storage capacity  $B=5$.
The user preference  is modeled by a five-state Markov chain with  different parameters $\alpha$ drawn from the kernel function.
The content popularity  is modeled by a four-state Markov chain drawn from Zipf distributions with different  distribution parameters $\beta$.
We choose the Least Recently Used (LRU) method, the Least Frequently Used (LFU)  method, and the  Q-learning method as the benchmark edge caching methods.

\begin{figure}[!t]
\centering 
\includegraphics[width=0.45 \textwidth]{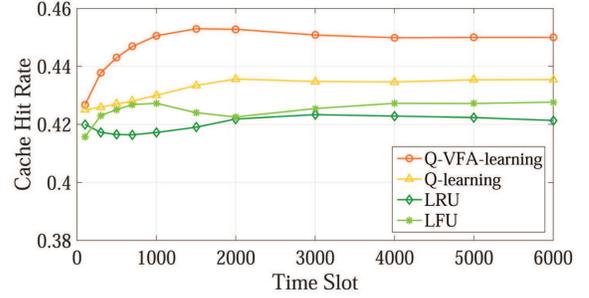}
\caption{Cache hit rate versus  time slot for the proposed  method and the benchmark methods.}
\label{chr}
\end{figure}

Fig. \ref{chr} shows the cache hit rate of our proposed method in comparison with the three benchmark  methods, and the instantaneous cache hit rate  is averaged  for every $100$ time slots to avoid contingency.
It can be observed that the performance of the Q-learning method and Q-VFA-learning method are superior to   those of the other methods, and the Q-VFA-learning method achieves the largest cache hit rate.
Moreover, the cache hit rates of the Q-learning method and the Q-VFA-learning method constantly increase until both reach the maximum, while the LRU and  LFU methods show fluctuations
at the beginning  as a result of the inevitable cold-start problem.

\begin{figure}[!t]
\centering
\includegraphics[width=0.45 \textwidth]{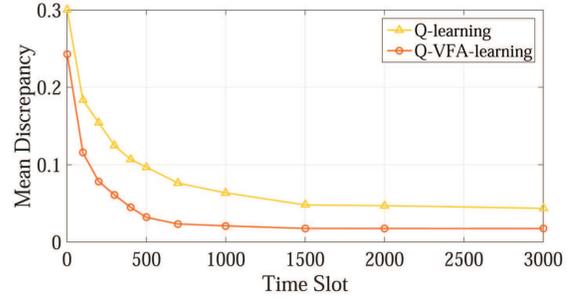}
\caption{Convergence rate of the Q-learning and the Q-VFA-learning.}
\label{conv}
\end{figure}

In Fig. \ref{conv}, we show the convergence performance of the Q-learning method and the Q-VFA-learning method in term of
the mean discrepancy, where the mean variance of  $Q$ values   is updated in every 100 iterations in order to avoid contingency.
We have set the discount factor $\gamma=0.9$ and the step size $\rho=0.005$ for a relatively fast convergence.
It can be observed that the Q-VFA-learning method converges faster than  the Q-learning method.
The reason is that the Q values in tabular form  must be updated per state-action pair, while Q-VFA-learning method reduces dimension of the problem as updating the learning parameters of  model characteristic per iteration.

\begin{figure}[!t]
\centering
\includegraphics[width=0.45 \textwidth]{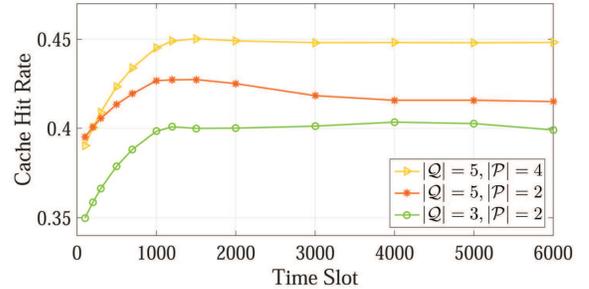}
\caption{Cache hit rate versus  time slot for the Q-VFA-learning  method under different dimensions of the content popularity Markov chain and the user preference Markov chain. }
\label{comp}
\end{figure}

In Fig. \ref{comp}, we compare the cache hit rate of the Q-VFA-learning method with different dimensions of the content popularity Markov chain and the user preference Markov chain.
It can be observed that  the  Q-VFA-learning method has better performance with  the increase of  $|\cal{P}|$ and $|\cal{Q}|$.
This reveals that the   our  proposed method can adapt to the complex environment, which contains the fluctuating content popularity and the dynamically arriving and leaving users.

\begin{figure}[!t]
\centering
\includegraphics[width=0.45 \textwidth]{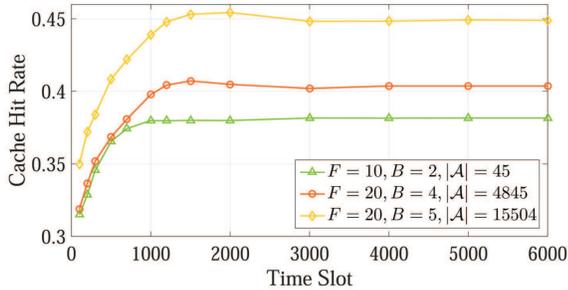}
\caption{Cache hit rate versus time slot for the Q-VFA-learning method under different   library size and the storage capacity of the F-AP.}
\label{sub_F}
\end{figure}

Fig. \ref{sub_F} shows the cache hit rate  of the Q-VFA-learning method under different library size and storage capacity of the F-AP.
It can be observed that with the increase of $|\cal{A}|$, the convergence becomes slower while the performance gets better.
This is due to the fact that the iteration process of the Q-VFA-learning method makes the edge caching policy  sensitive to the dimension of the action set $\cal{A}$.
With more explorations, the  F-AP can track and learn a large  content library more intelligently.
\section{Conclusions }
{\normalsize{
In this paper, we have proposed a RL-based distributed edge caching method  by learning and tracking the potential dynamic process of user requests.
The user request model based on hidden Markov process, which utilizes the joint influence of the local content popularity and the user preference, has been proposed to describe the fluctuant spatio-temporal traffic demands in F-RANs.
The Q-VFA-learning method has been proposed to seek the optimal edge caching policy and accelerate convergence in an online fashion.
Simulation results have shown that our proposed method   can achieve a better caching performance   compared to the  traditional methods.
}}

\section*{Acknowledgments}
{\small{
This work was supported in part by
the Natural Science Foundation of China under Grant 61521061,
the Natural Science Foundation of Jiangsu Province under grant
BK20181264,
the Research Fund of the State Key Laboratory of
Integrated Services Networks (Xidian University) under grant ISN19-10,
the Research Fund of the Key Laboratory of Wireless Sensor Network $\&$ Communication (Shanghai Institute of Microsystem and Information Technology, Chinese Academy of Sciences) under grant 2017002,
the National Basic Research Program of China
(973 Program) under grant 2012CB316004,
and the U.K. Engineering and Physical Sciences Research Council under Grant EP/K040685/2.
}}

\bibliographystyle{IEEEtran}      
\bibliography{ref}

\end{document}